%% file: arxiv.tex
\documentclass[10pt,twocolumn,letterpaper]{article}

\usepackage[pagenumbers]{cvpr} 


\usepackage{pifont}
\usepackage{graphicx}
\usepackage{amsmath}
\usepackage{amssymb}
\usepackage{booktabs}
\usepackage{xcolor}
\usepackage[english]{babel} 
\usepackage[T1]{fontenc}
\usepackage{array}
\usepackage{multirow}
\usepackage{amstext}
\PassOptionsToPackage{normalem}{ulem}
\usepackage{ulem}
\usepackage{bbding}
\usepackage{rotating}
\usepackage{colortbl}
\usepackage{enumitem}
\usepackage{tabularx}

\definecolor{cvprblue}{rgb}{0.21,0.49,0.74}
\usepackage[pagebackref,breaklinks,colorlinks,allcolors=cvprblue]{hyperref}

\title{
  \raisebox{-0.25\height}{\includegraphics[width=0.9cm]{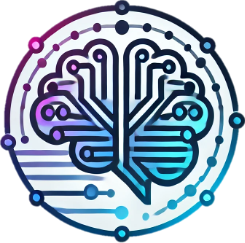}}
  CineBrain: A Large-Scale Multi-Modal Brain Dataset During Naturalistic Audiovisual Narrative Processing
}

\author{
Jianxiong Gao$^1$, Yichang Liu$^1$, Baofeng Yang$^1$, Jianfeng Feng$^1$, Yanwei Fu$^{1,2}$\\
$^1$Fudan University\quad $^2$Shanghai Innovation Institute\\
}

\begin{document}

\twocolumn[{
\renewcommand\twocolumn[1][]{#1}
\maketitle

\begin{center}
\vskip -0.05in
\includegraphics[width=1.0\linewidth]{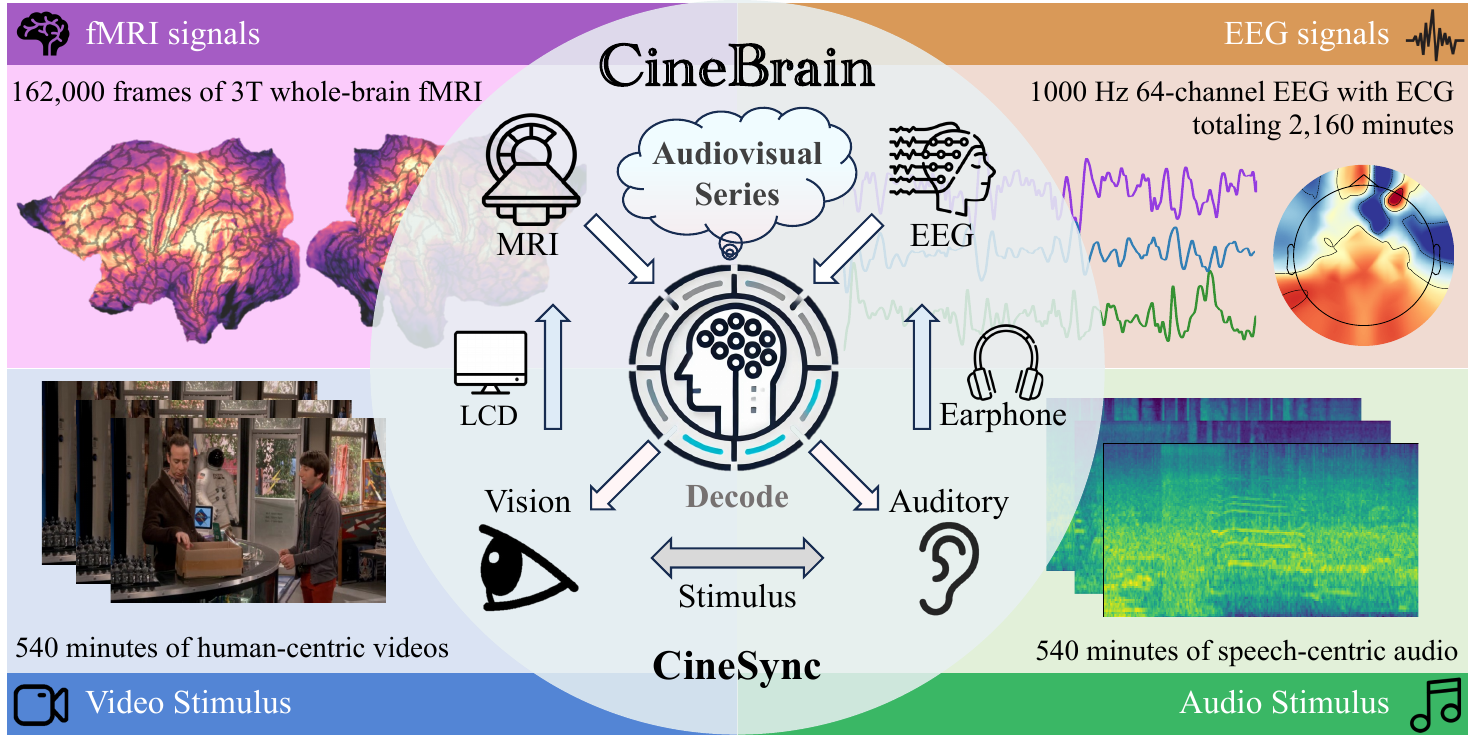}
\vskip -0.08in
\captionof{figure}{
\label{fig:teaser}
\textbf{Overview of CineBrain.} To leverage the complementary strengths of fMRI and EEG, CineBrain provides simultaneous audiovisual stimuli to participants while recording their EEG and fMRI signals. Engaging narrative-driven content from the television series \textit{The Big Bang Theory} is utilized to facilitate the study of complex brain dynamics and multimodal neural decoding.
}
\end{center}
}]

\begin{abstract}
Most research decoding brain signals into images, often using them as priors for generative models, has focused only on visual content. This overlooks the brain's natural ability to integrate auditory and visual information, for instance, sound strongly influences how we perceive visual scenes. To investigate this,
we propose a new task of reconstructing continuous video stimuli from multimodal brain signals recorded during audiovisual stimulation. To enable this, we introduce CineBrain, the first large-scale dataset that synchronizes fMRI and EEG during audiovisual viewing, featuring six hours of \textit{The Big Bang Theory} episodes for cross-modal alignment. We also conduct the first systematic exploration of combining fMRI and EEG for video reconstruction and present CineSync, a framework for reconstructing dynamic video using a Multi-Modal Fusion Encoder and a Neural Latent Decoder. CineSync achieves state-of-the-art performance in dynamic reconstruction, leveraging the complementary strengths of fMRI and EEG to improve visual fidelity. Our analysis shows that auditory cortical activations enhance decoding accuracy, highlighting the role of auditory input in visual perception. Project Page: \url{https://jianxgao.github.io/CineBrain}.
\end{abstract}

\section{Introduction}
\label{sec:intro}

In the vision community, considerable effort has been devoted to decoding brain signals—particularly fMRI—into images~\cite{chen2023seeing,mind-eye,scotti2024mindeye2sharedsubjectmodelsenable}, videos~\cite{mind-video,lichong_video}, and even 3D shapes~\cite{mind3d}, often using neural signals as priors to guide visual generative models. While this prior-guided approach is widely adopted, it largely reduces neural decoding to a tool for controlling generative models rather than probing the principles of human perception. In contrast, the human brain, arguably the most intricate biological system, excels at integrating information across multiple sensory modalities. Humans naturally perceive the world through rich audiovisual experiences, seamlessly combining visual and auditory cues to form coherent, dynamic representations of their surroundings. 
A classic example is the McGurk effect~\cite{mcgurk_nature}, where conflicting auditory and visual speech cues produce a third, illusory percept, highlighting the fundamental interplay between vision and hearing. Beyond generative priors, understanding how the brain achieves such  integration is central to bridging human cognition and machine perception.

Technically, most existing studies~\cite{chen2023seeing,mind-video,mind3d,eeg2video,eeg3d} in neural decoding focus on reconstructing visual content from visual-only stimuli, typically relying on a single neural modality such as fMRI or EEG. This limited scope overlooks two critical aspects of perception. First, auditory signals strongly modulate visual processing~\cite{sporns2007brain,park2013structural}, shaping attention, emotion, and context during naturalistic viewing. Second, fMRI and EEG provide complementary insights: fMRI offers fine-grained spatial localization, while EEG captures millisecond-level temporal dynamics. Ignoring the cross-modal and cross-modality interactions restricts the fidelity and realism of reconstructed visual stimuli.

\begin{figure}[t]
    \centering
    \includegraphics[width=\linewidth]{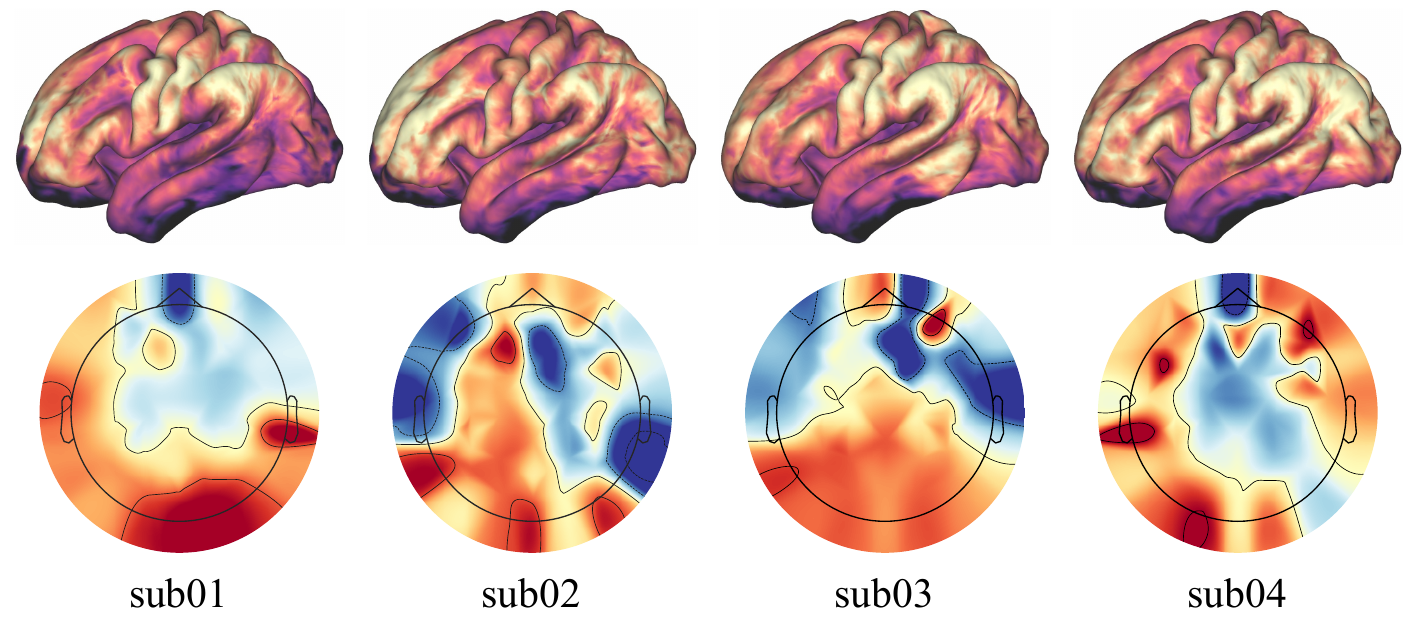}
\vskip -0.1in
\caption{
\textbf{Visualization of fMRI and EEG Responses in CineBrain.}
fMRI and EEG responses of subjects 1–4 to identical stimuli, illustrating individual differences in brain activation.
}
\vskip -0.1in
\label{fig:data_example}
\end{figure}

To address this gap, we propose both a new task and a novel large-scale dataset. Specifically, we present the task of reconstructing continuous video stimuli from multimodal brain signals recorded during naturalistic audiovisual stimulation. This task requires capturing both the spatial and temporal aspects of neural encoding, enabling exploration of how auditory input influences visual perception. However, progress in this area has been limited by the lack of datasets that synchronize fMRI and EEG recordings under ecologically valid, temporally dynamic audiovisual conditions. To overcome this challenge, we present \textbf{CineBrain}, the first large-scale multimodal dataset that synchronizes fMRI and EEG during naturalistic audiovisual viewing. Drawing inspiration from neurocinematics research~\cite{hasson2008neurocinematics,watabe2017neural}, which demonstrates that narrative-driven content naturally sustains attention and elicits complex brain dynamics, CineBrain includes approximately six hours of \textit{The Big Bang Theory} episodes, viewed by six participants under 3T fMRI scanning with concurrent EEG recordings captured using custom non-magnetic equipment. As illustrated in Fig.~\ref{fig:data_example}, the participants exhibit highly consistent activation patterns across both modalities, providing strong cross-modal alignment that is critical for multimodal decoding tasks.

Building on CineBrain, we conduct the first systematic exploration of fMRI–EEG integration for video reconstruction and present CineSync, a novel framework for reconstructing dynamic visual stimuli from multimodal brain signals. CineSync employs a Multi-Modal Fusion Encoder with a dual-transformer design to independently process fMRI and EEG sequences before merging them via a learned fusion projector. To achieve semantic alignment across modalities, brain representations are anchored to visual and textual embeddings using a joint contrastive objective. The resulting multimodal embeddings are decoded by a Neural Latent Decoder, a diffusion-based model that fuses brain-derived features with latent noise to generate semantically coherent and perceptually realistic videos.

Evaluation using semantic and perceptual metrics demonstrates that CineSync achieves state-of-the-art performance on temporally dynamic reconstruction within CineBrain. It effectively leverages the complementary strengths of fMRI and EEG to enhance visual fidelity. Notably, auditory-related cortical activations improve decoding accuracy, highlighting cross-modal facilitation in natural perception, and ablation studies show that increasing EEG representational capacity further boosts reconstruction quality, underscoring EEG’s critical role in capturing rapid neural dynamics for fine-grained video reconstruction.

In summary, our contributions are as follows:
\begin{itemize}
\item \textbf{A novel task}: 
We introduce the challenge of reconstructing continuous video stimuli from multimodal brain signals, modeling spatial and temporal neural encoding to reveal how auditory input influences visual perception.
\item  \textbf{A new dataset}: We present CineBrain, the first large-scale multimodal dataset that synchronizes fMRI and EEG during naturalistic audiovisual viewing, providing critical cross-modal alignment for decoding tasks.
\item  \textbf{A new framework}: We propose CineSync, a framework for reconstructing dynamic video stimuli using a dual-transformer-based Multi-Modal Fusion Encoder (MFE) and a diffusion-based Neural Latent Decoder.
\item  \textbf{State-of-the-art performance}: CineSync achieves superior reconstruction quality, leveraging fMRI–EEG complementarity, and highlighting the role of auditory-related cortical activations in improving decoding accuracy.
\end{itemize}

\section{Related Work}
\label{sec:relate_work}

\begin{table*}[htbp]
\small
    \centering
    \caption{
        \textbf{Overview of the CineBrain Dataset.} We present detailed statistics of our proposed CineBrain dataset and compare it with other existing video-based brain datasets. CineBrain provides comprehensive multimodal brain recordings during audiovisual stimulation. Each participant watched a total of 6 hours of audiovisual stimuli, corresponding to approximately 27,000 frames of fMRI data.
        \label{tab:detail_about_dataset}
    }
    \vskip -0.1in
    \setlength{\tabcolsep}{3mm}
    \resizebox{\linewidth}{!}{
    \begin{tabular}{lccccccc}
        \toprule
        \textbf{Dataset}  & \textbf{Stimulus Type}  & \textbf{Participants} & \textbf{Gender (M/F)} & \textbf{Duration} & \textbf{Videos} &  \textbf{EEG} & \textbf{fMRI} \\
        \midrule
        \midrule
        Wen~\cite{wen_video}                  & Video & 3    &  0/3 & 3.07 h  & 5520 & \ding{56} & \ding{51} \\
        fMRI-Video-FCVID~\cite{lichong_video} & Video & 3    &  1/1 & 1.11 h & 400 & \ding{56} & \ding{51}  \\
        fMRI-Video-WebVid~\cite{lichong_video}& Video & 5    &  2/2 & 2.67 h  & 1200 & \ding{56} & \ding{51} \\
        SEED-DV~\cite{SEED-DV} & Video & 20  & 10/10  & 0.76 h  & 1400 & \ding{51} & \ding{56}  \\
        \midrule
        \textbf{CineBrain (Ours)} & Audio+Video & 6  &  2/4 & 6 h & 5400 & \ding{51} & \ding{51}  \\
        \bottomrule
    \end{tabular}
    }
    \vskip -0.1in
\end{table*}

\subsection{Neuro-Stimulus Dataset}
Data is fundamental to deep learning. To facilitate neural decoding experiments, various datasets have been proposed. For visual stimuli,  NSD dataset~\cite{NSD} is a widely used large-scale fMRI dataset collected from image-based stimuli, while  Wen dataset~\cite{wen_video} and fMRI-Video-WebVid~\cite{lichong_video} contain fMRI data acquired from video stimuli. Additionally, the fMRI-Shape~\cite{mind3d} and fMRI-Objaverse~\cite{mind3d++} datasets provide fMRI recordings from 3D video stimuli. For EEG data, the SEED-DV~\cite{SEED-DV} dataset includes EEG recordings from video stimuli. Tang’s dataset~\cite{tang_audio} focuses specifically on auditory stimulation. However, these existing datasets are limited because each dataset only captures single-modality brain signals responding to single-modality stimuli. To address this limitation, in this paper, we propose a multimodal audiovisual stimulus dataset named CineBrain and conduct experiments based on it.

\subsection{Neural Decoding}
Neural decoding aims to reconstruct external stimuli perceived by the human brain from recorded neural signals. This task remains challenging due to the complex nature of both neural encoding and stimulus reconstruction.
Functional magnetic resonance imaging (fMRI), with its high spatial resolution and non-invasive acquisition, has shown strong performance in visual reconstruction. Recent works have achieved notable results in image reconstruction~\cite{chen2023seeing,mind-eye,sun2023contrastattenddiffusedecode} and extended these advances to video~\cite{mind-video,lichong_video,sun2025neuralflix,gong2024neuroclips,wang2025neuronsemulatinghumanvisual} and 3D reconstruction~\cite{mind3d,mind3d++}.
Electroencephalography (EEG), offering millisecond-level temporal resolution and continuous recording, has also proven effective for neural decoding. EEG-based methods have reconstructed images~\cite{singh2023eeg2imageimagereconstructioneeg,bai2023dreamdiffusiongeneratinghighqualityimages,song2024decoding}, videos~\cite{eeg2video,liu2025dynamindreconstructingdynamicvisual}, 3D structures~\cite{eeg3d}.
Despite these advances, most existing works have focused on single-modality decoding. The complementary characteristics of fMRI and EEG, where fMRI provides fine-grained spatial information and EEG offers precise temporal dynamics, have not yet been fully explored. In this paper, we investigate the integration of fMRI and EEG signals to exploit their complementary advantages and enhance the reconstruction of perceived stimuli.

\subsection{Diffusion Models}
Diffusion models~\cite{ddpm,ldm,sdxl} are a powerful generative framework known for their capability to produce high-quality images and can also be extended to other data modalities. The fundamental principle involves defining a forward diffusion process, in which clean data is progressively corrupted by adding Gaussian noise. A neural network, typically a U-Net architecture, is then trained to reverse this diffusion process, iteratively removing the noise to reconstruct high-quality data. DiT~\cite{DiT} has successfully scaled up diffusion models through transformer-based architectures, significantly improving their generative capabilities. Consequently, DiT has been effectively applied across multiple modalities, including text-to-image generation~\cite{DiT,sd3, flux2024}, text-to-video generation~\cite{cogvideox,he2022lvdm, chen2024videocrafter2}, and text-to-audio generation~\cite{f5tts, E2TTS}. In our study, motivated by the strong distribution modeling capabilities of diffusion models, we leverage diffusion models specifically trained on video and audio data to reconstruct these modalities from multimodal brain signals, including fMRI and EEG.

\section{Experimental Designs and Curated Dataset}
\label{sec:dataset}

In this section, we introduce CineBrain, the first multimodal brain dataset featuring synchronized EEG and fMRI recordings collected during story-based audiovisual stimulation. Tab.~\ref{tab:detail_about_dataset} summarizes key statistics of CineBrain and contrasts it with existing datasets primarily employing visual stimuli. In comparison, CineBrain provides richer data suitable for a broader range of downstream tasks. The dataset will be publicly available to promote research in story decoding and multimodal neural decoding.

\subsection{Subjects and Experiments}

Six participants (ages 21–26; two males, four females), who are unaware of the study’s objectives, contributed to the CineBrain dataset. All participants have normal or corrected-to-normal vision and hearing. Written informed consent was obtained, and the experimental protocol received approval from the appropriate ethics committee.

Since prolonged MRI scanning can cause discomfort and reduced attention, participants cannot stay in the scanner for long periods. Therefore, we select episodes from the popular TV series \textit{The Big Bang Theory} as audiovisual stimuli. The show’s engaging narratives, familiarity with daily life, and rich audiovisual content help maintain participants’ attention throughout the 18-minute viewing sessions. To ensure diverse yet comparable stimuli, participants collectively view 20 different episodes. All participants watch the first 10 episodes of Season 7; in addition, Participants 1, 2, and 6 view 10 episodes from Season 9, while the remaining participants watch 10 episodes from Season 11.

During the experiment, the video resolution is downsampled to 720p to match the MRI’s in-bore LCD screen. Episodes of varying lengths are standardized to 18 minutes of viewing time, easing subsequent data processing. Each 18-minute viewing constitutes one “run”, yielding 20 runs per participant. To maintain data quality and reduce fatigue, runs are grouped into sessions of two or three episodes each, with no more than five runs conducted on a single day. In total, each participant contributes approximately 6 hours of concurrent fMRI and EEG alongside ECG.

Visual stimuli are presented on an LCD screen ($8^\circ\times 8^\circ$) positioned at the head of the scanner bed. Participants viewed the screen through a mirror attached to the RF coil, maintaining fixation on a central red dot ($0.4^\circ\times 0.4^\circ$).

CineBrain experimental setup features simultaneous fMRI and EEG recording, harnessing their complementary strengths—high spatial resolution from fMRI and high temporal resolution from EEG. Given the noisy MRI environment, participants use sponge earplugs and custom-designed non-magnetic headphones with soft padding to ensure comfort and clear audio delivery. An MRI-compatible EEG cap is also worn, enabling simultaneous collection of these complementary neural data types, significantly broadening  dataset's potential for diverse downstream analyses.

\subsection{Data Acquisition and Preprocessing}

\noindent \textbf{fMRI.} 
The fMRI acquisition protocol follows established experimental standards~\cite{mind3d,mind3d++,lichong_video}. A 3T scanner equipped with a 32-channel RF head coil is employed to obtain high-resolution T1-weighted structural images and functional data for the decoding experiments.
T1-weighted images were obtained using an MPRAGE sequence (0.8-mm isotropic resolution, TR = 2500 ms, TE = 2.22 ms, flip angle $8^{\circ}$). Functional data were recorded using a gradient-echo EPI sequence with whole-brain coverage (2-mm isotropic resolution, TR = 800 ms, TE = 37 ms, flip angle $52^{\circ}$, multi-band factor = 8). With a sampling frequency of 1.25 Hz, each video run generated 1350 functional MRI frames.

\begin{figure}[t]
    \centering
    \includegraphics[width=\linewidth]{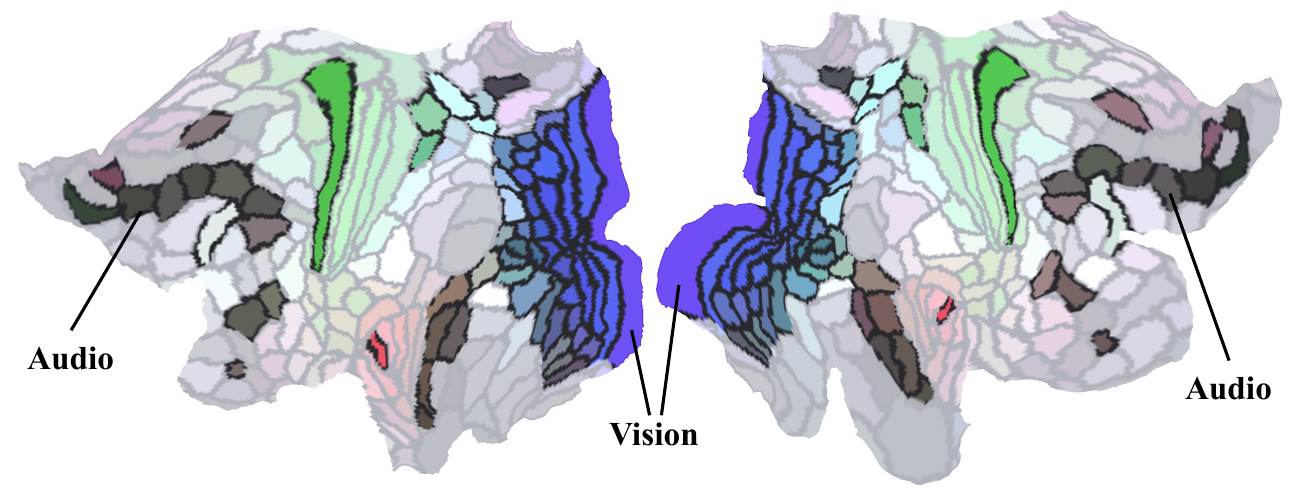}
    \vskip -0.1in
\caption{
ROIs from the fMRI signals used in our experiments.
\label{fig:fmri_roi_vis}
}
\vskip -0.15in
\end{figure}

Preprocessing is performed using the widely adopted fMRIPrep pipeline~\cite{fmriprep1,fmriprep2}.
Unlike previous studies that focused exclusively on visual decoding~\cite{mind3d, chen2023seeing, mind-video}, our experimental design additionally incorporates auditory stimuli, prompting the selection of both visual and auditory regions of interest (ROIs).
The selected ROIs are illustrated in Fig.~\ref{fig:fmri_roi_vis}.
Specifically, the visual ROIs comprise 8,405 voxels, while the auditory ROIs comprise 10,541 voxels.
Detailed ROI definitions are provided in the supplementary material.

\begin{figure*}
    \centering
    \includegraphics[width=\linewidth]{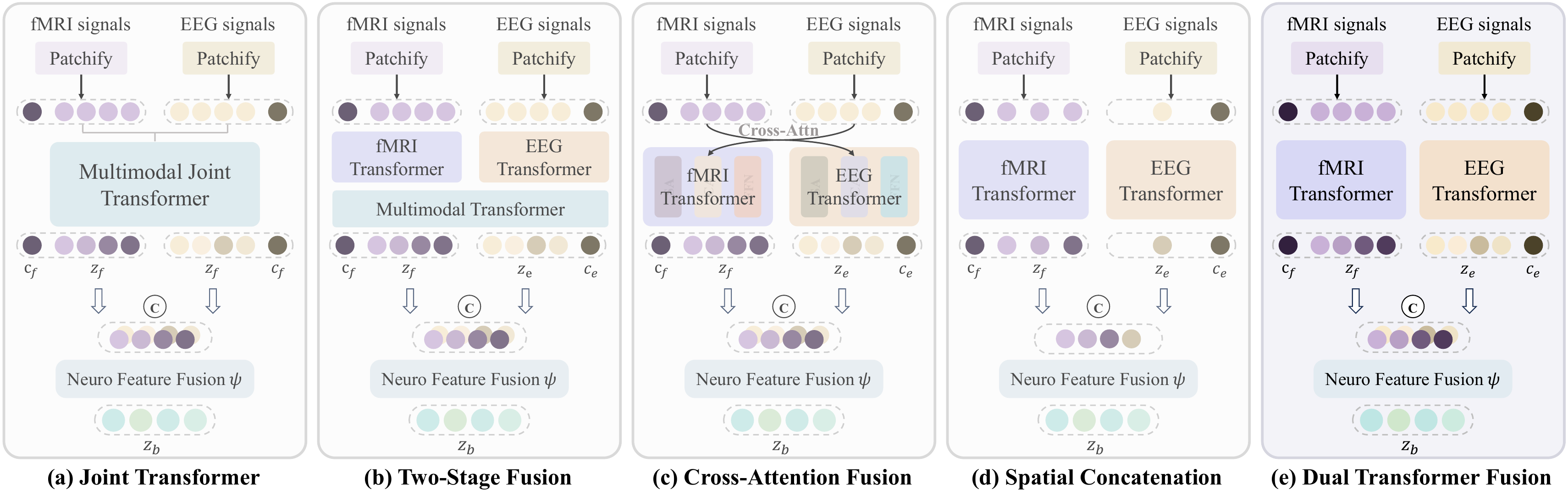}
    \vskip -0.1in
\caption{
\textbf{Architectural exploration for integrating fMRI and EEG.} 
We compare five encoder variants, each adopting a distinct fusion mechanism to integrate the complementary information from fMRI and EEG signals.
}
\label{fig:arch_discuss}
\vskip -0.1in
\end{figure*}

To account for the inherent delay in BOLD responses, fMRI signals underwent z-scoring across vertices within each run, incorporating a 4-second lag.

\noindent \textbf{EEG.} EEG data are captured using an MRI-compatible 64-channel cap at 1000 Hz, simultaneously recording ECG signals. Precise synchronization between EEG and fMRI data is ensured by logging fMRI TR timings. EEG preprocessing involves a multi-step artifact removal approach, targeting scanner-induced noise and biological interference while preserving neural activity. This includes bandpass filtering (0.1–30 Hz) to remove baseline drift and muscle artifacts, and a 50 Hz notch filter to reduce powerline interference~\cite{chen2022dataset, lee2024eav, li2024visual, li2025neurobolt}. ECG artifacts are initially mitigated using QRS-based techniques, followed by independent component analysis to isolate residual artifacts. 
ECG recordings further support adaptive refinement of artifact removal, yielding clean EEG data for subsequent analysis.

\noindent \textbf{Video Stimuli.}
The 24 fps, 720p video stimuli consist of 20 episodes, each lasting 18 minutes. For video decoding, we downsample the videos to a resolution of 480 × 720 and segment them into 4-second clips (33 frames per clip). Each participant thus contributes 5,400 clips in total, including 4,860 for training (from the first 18 episodes) and 540 for testing (from the final two episodes).

\noindent \textbf{Audio Stimuli.}
Audio from each video episode is also segmented into corresponding 4-second clips, resulting in the same number of samples as the video clips, maintaining consistency for training and evaluation.

\noindent \textbf{Text Descriptions.}
To support multimodal decoding experiments, textual descriptions are generated separately for each audio and video clip. Video descriptions are generated using Qwen2.5-VL~\cite{Qwen2.5-VL} to facilitate contrastive learning. Audio clips are transcribed using Whisper-large-v3~\cite{whisper}, providing rich textual inputs aligned with auditory data for enhanced multimodal analyses.

\subsection{Extended Downstream Tasks}
Although CineBrain is primarily collected to support multimodal brain-signal-based video decoding, it also provides a solid foundation for a range of extended downstream tasks that may inspire broader multimodal research:
\textbf{(1)} \textit{Auditory decoding}: leveraging both fMRI and EEG signals to reconstruct auditory stimuli.
\textbf{(2)} \textit{EEG-to-fMRI translation}: exploring cross-modal mappings from EEG to fMRI representations, which is particularly meaningful given the high cost and difficulty of fMRI acquisition.
\textbf{(3)} \textit{Stimulus-to-brain modeling}: further extending the framework to predict fMRI or EEG responses directly from video or audio stimuli.
Together, these tasks demonstrate the extensibility of CineBrain beyond video reconstruction. We envision CineBrain serving as a comprehensive benchmark for advancing research on how the human brain perceives, integrates, and reconstructs dynamic audiovisual experiences.

\begin{figure*}[t]
    \centering
    \includegraphics[width=\linewidth]{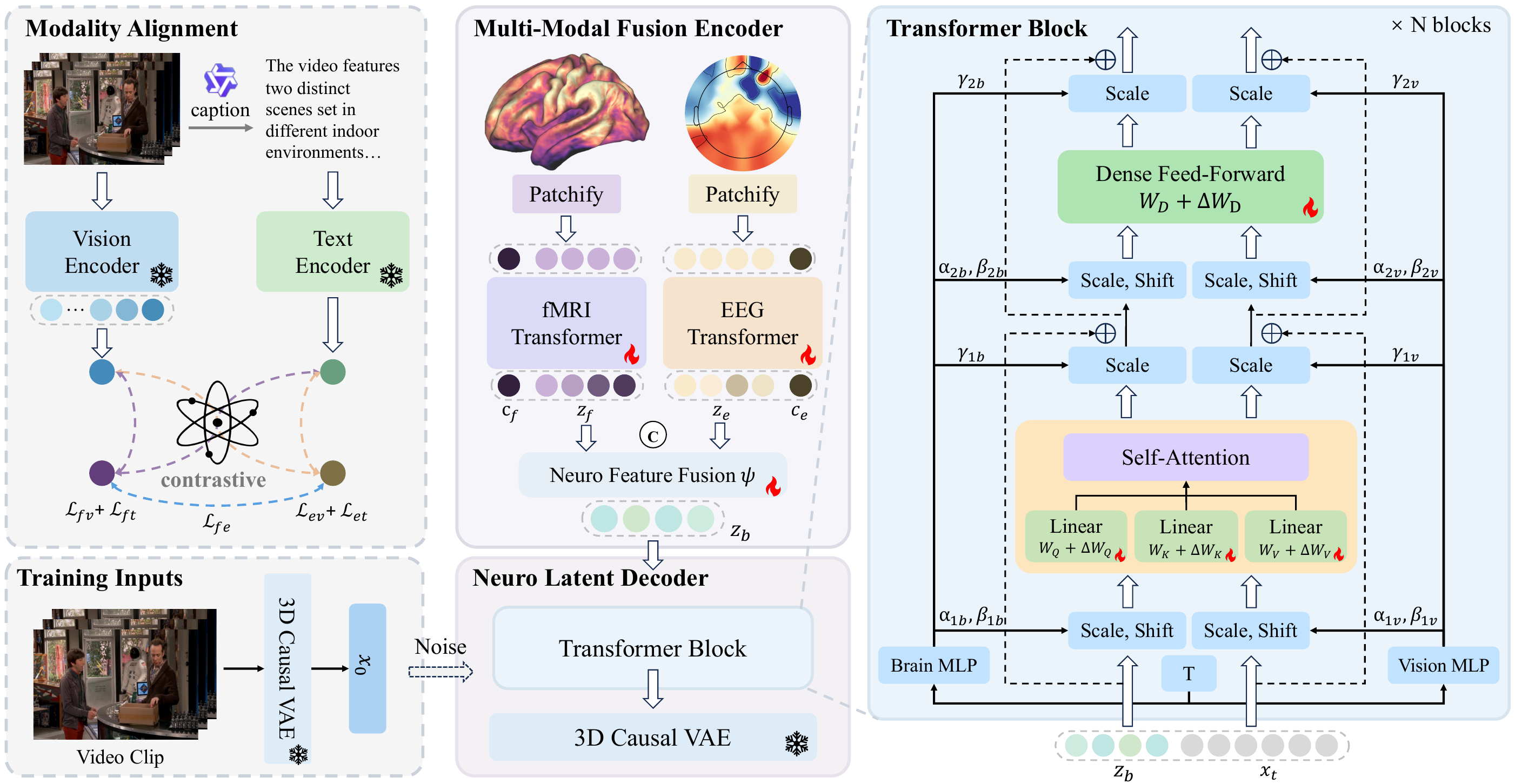}
\vskip -0.1in
\caption{
\textbf{Overview of the CineSync Framework.}
CineSync first employs a Multimodal Fusion Encoder to extract features from fMRI and EEG data, with a modality alignment module to align these features with semantic information. Subsequently, it utilizes a LoRA-tuned neural latent decoder to reconstruct videos based on the fused brain features. \textit{Note: The gray box is used only during training.}
\label{fig:framework}
}
\vskip -0.1in
\end{figure*}

\section{Methods}
\label{sec:method}

\noindent \textbf{Overall.}
To evaluate the effectiveness of our proposed CineBrain framework and the integration of fMRI and EEG signals to enhance downstream decoding tasks, we introduce an innovative and versatile approach comprising two main components:
\textbf{1) Multi-Modal Fusion Encoder (MFE)}: This module extracts semantically aligned features from fMRI and EEG signals and subsequently fuses these modalities to produce effective representations for downstream tasks. \textbf{2) Neuro Latent Decoder (NLD)}: Utilizing fused signals from the MFE, this module employs a LoRA-tuned, diffusion-based decoder to reconstruct corresponding stimuli (e.g., video or audio) from brain signals.

\begin{table}[tb]
\small
\centering{
\caption{
\label{tab:encoder_compare}
\textbf{Performance comparison of different multimodal encoder structures.} We evaluate all variants using video-level semantic metrics and frame-level perceptual metrics.
\textbf{Bold} denotes the best performance, while \underline{underlined} denotes the second-best.
}
\vskip -0.1in
\resizebox{\linewidth}{!}{
\begin{tabular}{l|ccc|cc}
\toprule
\multicolumn{1}{c|}{\multirow{2}{*}{\textsc{Methods}}} & 
\multicolumn{3}{c|}{Semantic-level} & 
\multicolumn{2}{c}{Perceptual-level} \\
 & 2-way$\uparrow$ & 50-way$\uparrow$ & \multicolumn{1}{c|}{FVD$\downarrow$} & SSIM$\uparrow$ & PSNR$\uparrow$ \\
\midrule
\midrule
(a) Joint Transformer            & 0.924 & 0.274 & \multicolumn{1}{c|}{128.0}  & 0.232 & 9.30 \\
(b) Two-Stage Fusion        & 0.921 & 0.278 & \multicolumn{1}{c|}{120.0}  & 0.242 & 9.60 \\
(c) Cross-Attn. Fusion      & 0.921 & 0.292 & \multicolumn{1}{c|}{110.0}  & 0.249 & 9.90 \\
\midrule
(d) SpatialCat (f113-e113) & 0.918 & 0.276 & \multicolumn{1}{c|}{114.0}  & 0.255 & 9.95 \\
(d) \underline{SpatialCat (f34-e192)}  & 0.928 & 0.311 & \multicolumn{1}{c|}{106.0}  & 0.250 & 10.10 \\
(d) SpatialCat (f24-e202)  & 0.926 & 0.307 & \multicolumn{1}{c|}{108.0}  & 0.247 & 10.00 \\
\midrule
\textbf{(e) Dual Trans. Fusion}  & 0.929 & 0.324 & \multicolumn{1}{c|}{51.53} & 0.249 & 12.03 \\
\bottomrule
\end{tabular}}
}
\vskip -0.15in
\end{table}

\subsection{Multi-Modal Fusion Encoder}


\noindent \textbf{fMRI–EEG Fusion Exploration.}
As an initial exploration of fMRI–EEG integration, we evaluate five fusion architectures under comparable parameter budgets, as illustrated in Fig.~\ref{fig:arch_discuss}.
(a) \textbf{Joint Transformer}: concatenates fMRI and EEG features and feeds them into a unified Transformer for joint representation learning.
(b) \textbf{Two-Stage Fusion}: processes fMRI and EEG separately with individual Transformers, followed by a joint Transformer for integration.
(c) \textbf{Cross-Attention Fusion}: employs modality-specific Transformers where each block includes cross-attention for inter-modal information exchange.
(d) \textbf{Spatial Concatenation}: uses different token counts for fMRI and EEG features, doubles the feature dimension, and spatially concatenates them before fusion. For a fair comparison, the total token count is halved.
(e) \textbf{Dual Transformer Fusion}: extracts features from each modality with independent Transformers and aggregates them at the final stage. 
Tab.~\ref{tab:encoder_compare} indicates that separately encoding fMRI and EEG signals yields better performance.
This finding reveals a substantial discrepancy between the two modalities, suggesting that directly applying fully shared self-attention is not suitable for their fusion. It also provides important insights that inform the design of our subsequent fusion module.

Motivated by this observation, we design a dual-transformer architecture termed the \textbf{Multi-Modal Fusion Encoder (MFE)}. Inspired by the success of contrastive learning~\cite{clip} in neuro-decoding~\cite{mind-video, lichong_video, mind3d++} and multi-modal representation learning, the MFE adopts the Vision Transformer (ViT) as the backbone with a class token to semantically align information across modalities.
Formally, we denote the fMRI and EEG signals as $\mathbf{x}_f$ and $\mathbf{x}_e$, respectively.  
The MFE is defined as $E = \{E_f, E_e\}$, where $E_f$ and $E_e$ are modality-specific encoders for fMRI and EEG.  
Through the MFE, we obtain both latent features and class tokens:
\begin{equation}
\mathbf{z}_f, \mathbf{z}_e, \mathbf{c}_f, \mathbf{c}_e = E(\mathbf{x}_f, \mathbf{x}_e),
\end{equation}
where $\mathbf{z}_f$ and $\mathbf{z}_e$ denote the latent representations extracted from fMRI and EEG, and $\mathbf{c}_f$ and $\mathbf{c}_e$ are their corresponding class tokens used for contrastive alignment.

To integrate information across modalities, we introduce a fusion MLP $\psi$ that combines the latent features:
\begin{equation}
\mathbf{z}_b = \psi(\mathbf{z}_f, \mathbf{z}_e),
\end{equation}
where $\mathbf{z}_b$ serves as the unified brain representation for subsequent video reconstruction.

For cross-modal alignment, we employ pre-trained contrastive encoders $E_v$ and $E_t$ to extract embeddings from videos and their textual descriptions generated by the VLM (see Fig.~\ref{fig:framework}).  
Given a video clip with $n$ frames $\mathcal{V} = \{I_1, \dots, I_n\}$, frame-wise embeddings are obtained via $E_v$ and aggregated by a temporal aggregation module $\varphi$ to form a video-level representation $\mathbf{c}_v$.  
For text, we directly encode the caption using $E_t$:
\begin{equation}
\mathbf{c}_v = \varphi\bigl(\{E_v(I_i)\}_{i=1}^n\bigr), 
\quad 
\mathbf{c}_t = E_t(\text{Text}).
\end{equation}
We define the contrastive objectives as:
\begin{equation}
\begin{aligned}
\mathcal{L}_{fv} &= \mathcal{L}_{\text{clip}}(\mathbf{c}_f, \mathbf{c}_v), \quad
\mathcal{L}_{ft} = \mathcal{L}_{\text{clip}}(\mathbf{c}_f, \mathbf{c}_t), \\
\mathcal{L}_{ev} &= \mathcal{L}_{\text{clip}}(\mathbf{c}_e, \mathbf{c}_v), \quad
\mathcal{L}_{et} = \mathcal{L}_{\text{clip}}(\mathbf{c}_e, \mathbf{c}_t).
\end{aligned}
\end{equation}
To further align fMRI and EEG representations, we introduce an additional cross-modal contrastive loss:
\begin{equation}
\mathcal{L}_{fe} = \mathcal{L}_{\text{clip}}(\mathbf{c}_f, \mathbf{c}_e).
\end{equation}
Thus, the overall contrastive objective is given by:
\begin{equation}
\mathcal{L}_{\text{c}} = \mathcal{L}_{fv} + \mathcal{L}_{ft} + \mathcal{L}_{ev} + \mathcal{L}_{et} + \mathcal{L}_{fe}.
\end{equation}
During training, we optimize the MFE, the fusion MLP $\psi$, and the aggregation module $\varphi$, while keeping the pre-trained encoders $E_v$ and $E_t$ frozen.  
After extracting the fused brain representation $\mathbf{z}_b$, it is fed into the subsequent module for video reconstruction.

\begin{figure*}[t]
\centering
\includegraphics[width=\linewidth]{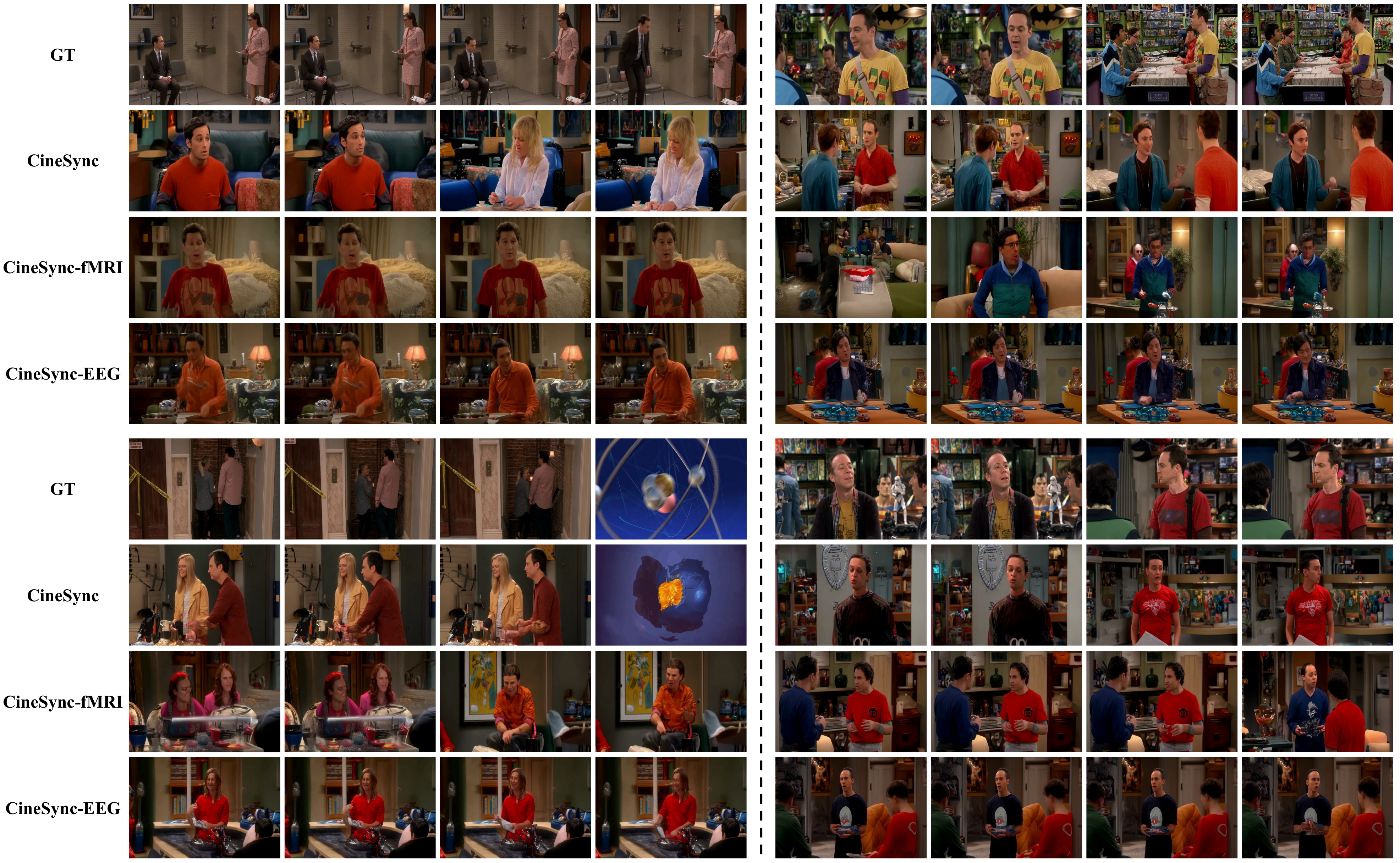}
\vskip -0.1in
\caption{
\textbf{Qualitative comparison of our method with baselines.} We compare the results of CineSync, CineSync-fMRI, and CineSync-EEG with the ground truth (GT). CineSync demonstrates higher accuracy, greater temporal consistency, and improved video quality.
}
\vskip -0.1in
\label{fig:video_res}
\end{figure*}

\subsection{Neuro Latent Decoder}
To effectively reconstruct the temporal stimuli, we design a \textbf{Neuro Latent Decoder (NLD)}, a general-purpose module that can be seamlessly integrated into any video diffusion model. Specifically, for video reconstruction, we adopt and adapt CogVideoX-5B~\cite{cogvideox} as the underlying diffusion model within our NLD. CogVideoX-5B generates videos at 8~fps with a resolution of $480\times720$, conditioned on text prompts. Given the central role of textual embeddings in video diffusion models, we replace the original text condition with our fused brain representation $\mathbf{z}_b$. 

During training, a video clip $V$ is first encoded into the latent space using a 3D-VAE $\mathcal{E}$, after which noise is added according to the forward diffusion process:
\begin{equation}
\mathbf{x}_0 = \mathcal{E}(V), \quad 
\mathbf{x}_t = \sqrt{\overline{\alpha}_t}\,\mathbf{x}_0 + \sqrt{1 - \overline{\alpha}_t}\,\boldsymbol{\epsilon}, 
\quad
\boldsymbol{\epsilon} \sim \mathcal{N}(\mathbf{0}, \mathbf{I}),
\end{equation}
where $t$ is uniformly sampled from $\{1, \dots, 1000\}$. Following the training strategy of CogVideoX, we adopt explicit uniform timestep sampling. As illustrated in Fig.~\ref{fig:framework}, the noised latent $\mathbf{x}_t$ is concatenated with the fused brain feature $\mathbf{z}_b$ and then fed into the diffusion model. 

To efficiently adapt the model to brain-derived inputs, we employ LoRA fine-tuning on both the attention and feed-forward layers within the DiT blocks of NLD, enabling effective integration of neural representations into the video generation process.
The training objective of the NLD follows the standard diffusion loss:
\begin{equation}
\mathcal{L} = 
\mathbb{E}_{V,\, \boldsymbol{\epsilon},\, t}
\left[
\left\|
\boldsymbol{\epsilon} - \boldsymbol{\epsilon}_\theta(\mathbf{x}_t,\, \mathbf{z}_b,\, t)
\right\|^2
\right],
\quad 
\boldsymbol{\epsilon} \sim \mathcal{N}(\mathbf{0}, \mathbf{I}).
\end{equation}

\begin{table*}[tb]
\small
\centering{
\caption{
\label{tab:video_res_compare}
\textbf{Performance comparison of CineSync with baselines.} The average metrics across all subjects are reported. CineSync$^{\star}$ indicates the experiment that includes audio-related ROIs in fMRI. \textbf{Bold} denotes the best performance, while \underline{underlined} denotes the second-best.
}
\vskip -0.1in
    \setlength{\tabcolsep}{3.4mm}{
    \begin{tabular}{l|ccccc|cccc}
    \toprule
    \multicolumn{1}{c|}{\multirow{3}{*}{\textsc{Methods}}} & \multicolumn{5}{c|}{Semantic-level} & \multicolumn{4}{c}{Perceptual-level} \\
    & \multicolumn{3}{c|}{Video-based} &  \multicolumn{2}{c|}{Frame-based} &  \multicolumn{2}{c|}{Video-based} &  \multicolumn{2}{c}{Frame-based} \\
     & 2-way$\uparrow$ & 50-way$\uparrow$ & \multicolumn{1}{c|}{FVD$\downarrow$} & 2-way$\uparrow$  & 50-way$\uparrow$ & DTC$\uparrow$ & \multicolumn{1}{c|}{CTC$\uparrow$}  & SSIM$\uparrow$ & PSNR$\uparrow$ \\
    \midrule
    \midrule

EEG2Video~\cite{eeg2video} & 0.786  & 0.162  & \multicolumn{1}{c|}{146.23} & 0.815 & 0.134 & 0.683 & 0.707  & 0.109 & 6.218  \\
CineSync-EEG   & 0.891  & 0.304  & \multicolumn{1}{c|}{53.75}  & 0.918 & 0.349 & 0.899 & 0.937 & 0.231   & 11.75 \\ 
\midrule
GLFA~\cite{lichong_video} & 0.801 & 0.167 & \multicolumn{1}{c|}{128.76} & 0.847 & 0.225 & 0.706 & 0.735  & 0.123 &  7.526  \\
NeuroClips~\cite{gong2024neuroclips} & 0.816  & 0.183  & \multicolumn{1}{c|}{116.36} & 0.833 & 0.142 & 0.871 & 0.868  & 0.087 & 6.854 \\
CineSync-fMRI  & 0.893  & 0.307  & \multicolumn{1}{c|}{57.47} & 0.926 & 0.358 & 0.907 & 0.945  & 0.240   & 11.92 \\ 
\midrule
CineSync &   \underline{0.909}  & \underline{0.319}  & \multicolumn{1}{c|}{\underline{52.78}}  &  \underline{0.937} & \underline{0.398} & \underline{0.915} & \textbf{0.967} & \underline{0.262} & \underline{11.99}  \\ 
\textbf{CineSync$^{\star}$} &   \textbf{0.926}  & \textbf{0.336}  & \multicolumn{1}{c|}{\textbf{44.77}}  &  \textbf{0.954} & \textbf{0.423} & \textbf{0.921} & \underline{0.953} & \textbf{0.297} & \textbf{12.18}  \\ 
    \bottomrule
    \end{tabular}}
    }
\vskip -0.05in
\end{table*}

\section{Experiments}
\label{sec:exp}

\noindent \textbf{Metrics.}  
To comprehensively evaluate the model’s performance, we assess the reconstructed videos at two levels: semantic and perceptual.
\textbf{1) Semantic-Level:} Semantic similarity is essential for evaluating reconstruction quality. Following prior neuro-decoding studies~\cite{lichong_video,mind-video,mind-eye,mind3d,eeg2speech}, we compute N-way top-K accuracy across entire videos and individual frames to measure feature similarity. Additionally, the Fréchet Video Distance (FVD) is calculated to quantify the distributional similarity between reconstructed and ground-truth videos.
\textbf{2) Perceptual-Level:} Beyond semantics, visual quality is also critical. We employ DINO temporal consistency (DTC)~\cite{oquab2023dinov2} and CLIP temporal consistency (CTC)~\cite{clip} to assess video temporal coherence. Structural similarity (SSIM) and peak signal-to-noise ratio (PSNR) are further used to evaluate frame-level visual fidelity.

\subsection{Implementation Details}
Both the fMRI and EEG transformers contain 12 layers with a hidden dimension of 2048 and a token length of 227 (226 spatial tokens plus one class token).
The LoRA configuration in the NLD adopts a rank of 64 and a scaling factor $\alpha=64$.
For modality alignment, we employ SigLIP~\cite{siglip} to extract visual and textual embeddings.
Each input sample comprises 4-second multimodal brain signals, corresponding to $5\times8{,}405$ fMRI voxels and $64\times4{,}000$ EEG data points.
We use the AdamW optimizer with $\beta=(0.9, 0.95)$ and an initial learning rate of $1\times10^{-4}$.
Further implementation details are provided in the supplementary material.

\subsection{Experimental Results}
\noindent \textbf{Encoder Comparison.}  
As shown in Tab.~\ref{tab:encoder_compare}, we observe that reducing the interaction between fMRI and EEG during the feature extraction stage leads to better overall performance.
This finding suggests that the two modalities exhibit substantial representational differences, and excessive early fusion may hinder effective feature learning.
Interestingly, experiment (d) shows that when the total token count is fixed at 226, moderately increasing the number of EEG tokens leads to performance gains in video reconstruction.
However, we also observe that although this design expands the feature dimension, the reduced overall token count yields suboptimal performance compared with method (e), indicating that the number of tokens plays a more critical role.

\noindent \textbf{Comparison with Baselines.}  
To evaluate the effectiveness of our proposed CineSync, we compare it with several representative baselines, including EEG2Video~\cite{eeg2video}, GLFA~\cite{lichong_video}, and NeuroClips~\cite{gong2024neuroclips}, as well as two simplified variants of our model: CineSync-fMRI and CineSync-EEG.
All models are trained and evaluated on the same dataset for fair comparison.
Tab.~\ref{tab:video_res_compare} reports the average performance across all subjects.
Overall, the two simplified versions of CineSync substantially outperform the baselines across all metrics, validating the robustness of our framework even when using a single modality.
Furthermore, the full CineSync model surpasses both CineSync-fMRI and CineSync-EEG, demonstrating that jointly modeling fMRI and EEG signals enables more comprehensive decoding of brain activity and leads to higher-quality video reconstruction.
These results establish CineSync as a new state-of-the-art framework for reconstructing dynamic visual stimuli from human brain signals.
In addition, we conduct an extended experiment by incorporating auditory ROIs into the fMRI inputs for video reconstruction, while retaining all EEG data, resulting in a total of 18,946 voxels per frame.
We refer to this variant as CineSync$^{\star}$.
The results show that incorporating auditory ROIs improves video reconstruction performance, underscoring the benefits of multimodal and cross-regional integration in brain decoding.

\noindent \textbf{Qualitative Analysis.} 
Qualitative results in Fig.~\ref{fig:video_res} confirm that CineSync captures finer temporal dynamics and richer perceptual details than the baselines.
This suggests that our framework effectively leverages the complementary advantages of fMRI and EEG to enhance reconstruction fidelity.
Moreover, the consistently high reconstruction quality across subjects highlights the reliability and value of the CineBrain dataset in supporting 
multimodal brain-signal research, potentially facilitating further progress at the intersection of computer vision and cognitive neuroscience.

\noindent \textbf{Ablation Studies.}
We conduct an ablation study to evaluate the effectiveness of the contrastive learning–based alignment employed in our framework. 
The results, presented in the supplementary material, demonstrate that our multi-level contrastive alignment strategy significantly contributes to the overall performance, confirming its effectiveness in aligning multimodal brain representations.

\section{Conclusion}
\label{sec:exp}
In this paper, we present CineBrain, the first large-scale multimodal dataset simultaneously recording fMRI and EEG under naturalistic audiovisual stimulation.
Building on this dataset, we introduce the task of reconstructing continuous video stimuli from multimodal brain signals.
Furthermore, we explore effective strategies for integrating fMRI and EEG and propose CineSync, an innovative framework that integrates fMRI and EEG to enhance video reconstruction quality.
CineSync achieves state-of-the-art performance, consistently outperforming baselines using single-modal input, thereby demonstrating the benefits of multimodal brain representations.
Experiments further show that incorporating auditory ROIs improves visual decoding, 
reinforcing the reliability and value of CineBrain dataset and highlighting its potential to advance multimodal brain decoding research.
We hope our dataset and framework will serve as strong baselines and inspire future progress at the intersection of computer vision and cognitive neuroscience.

{
    \small
    \bibliographystyle{ieeenat_fullname}
    \bibliography{main}
}

\clearpage
\setcounter{page}{1}
\maketitlesupplementary

\section{Detailed Information on CineBrain}
\label{sec:dataset}
\subsection{Detailed ROIs Selected in the Experiments}
The regions of interest (ROIs) used in our study are summarized below. For the visual cortex, we include the following ROIs:
\textit{V1, V2, V3, V3A, V3B, V3CD, V4, LO1, LO2, LO3, PIT, V4t, V6, V6A, V7, V8, PH, FFC, IP0, MT, MST, FST, VVC, VMV1, VMV2, VMV3, PHA1, PHA2, PHA3}.
For the auditory-related cortex, we select ROIs corresponding to:
\textit{4, 7AL, 7Am, 7m, 7PC, 7PL, 7Pm, 8Ad, 8Av, 8BM, 8C, 9a, 9p, 10d, 10v, 44, 45, 46, 47m, A1, IPS1, p32, PGp, s32, STGa, STSda, STSdp, STSva, STSvp, TPOJ1, TPOJ2, TPOJ3}. In total, the selected visual ROIs include 8,405 voxels, while the auditory ROIs include 10,541 voxels.

\subsection{Prompts for Generating Text Descriptions}
\begin{figure}
\centering
\includegraphics[width=1\linewidth]{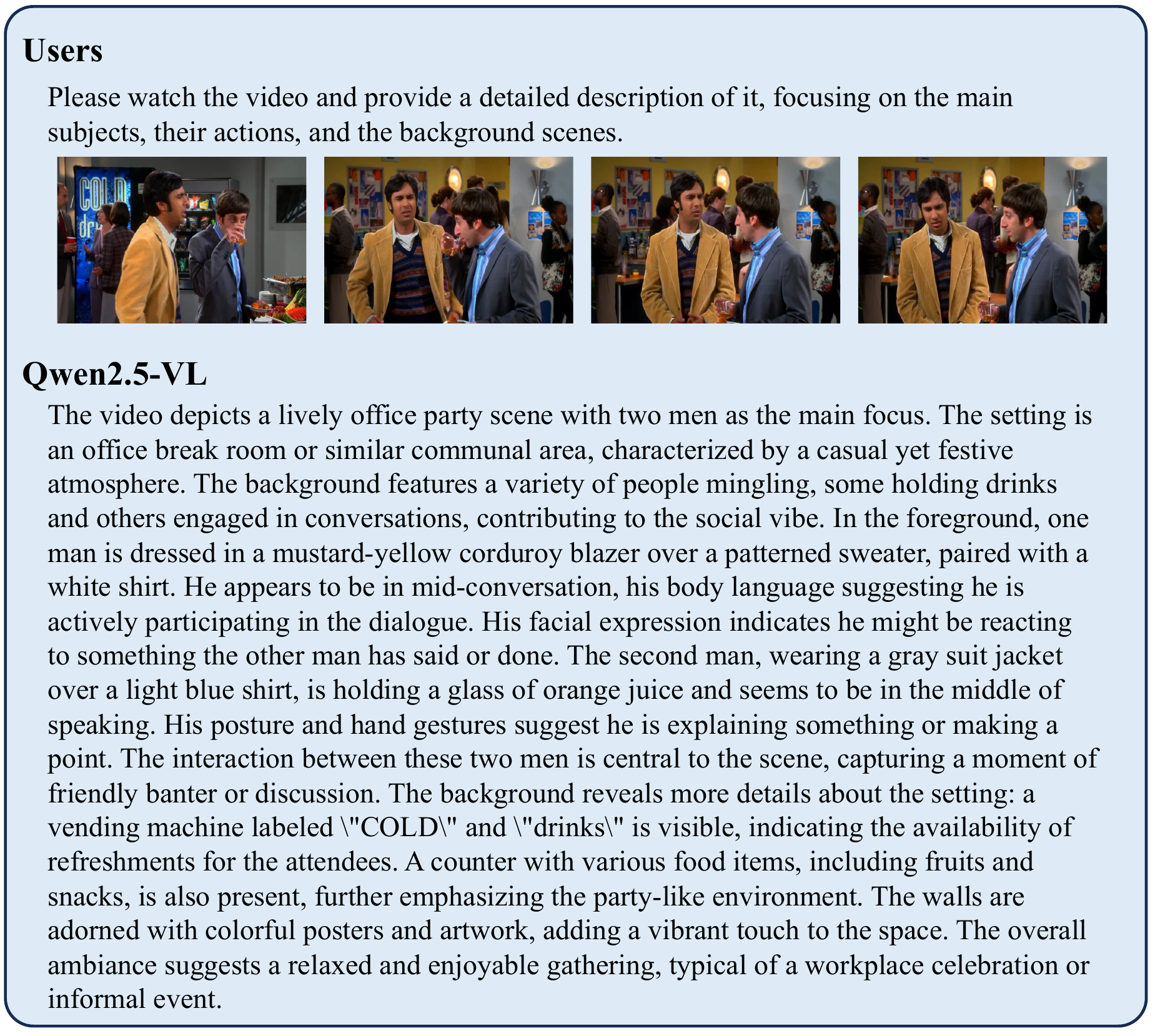}
\vskip -0.1in
\caption{
Prompt utilized for generating video descriptions.
\label{fig:qwen}
}
\vskip -0.1in
\end{figure}

Textual descriptions play an important role in our training pipeline. The prompts used to generate these descriptions with Qwen2.5-VL~\cite{Qwen2.5-VL} are provided in Fig.~\ref{fig:qwen}.

\subsection{EEG Experimental Device}
\begin{figure}
\centering
\includegraphics[width=0.8\linewidth]{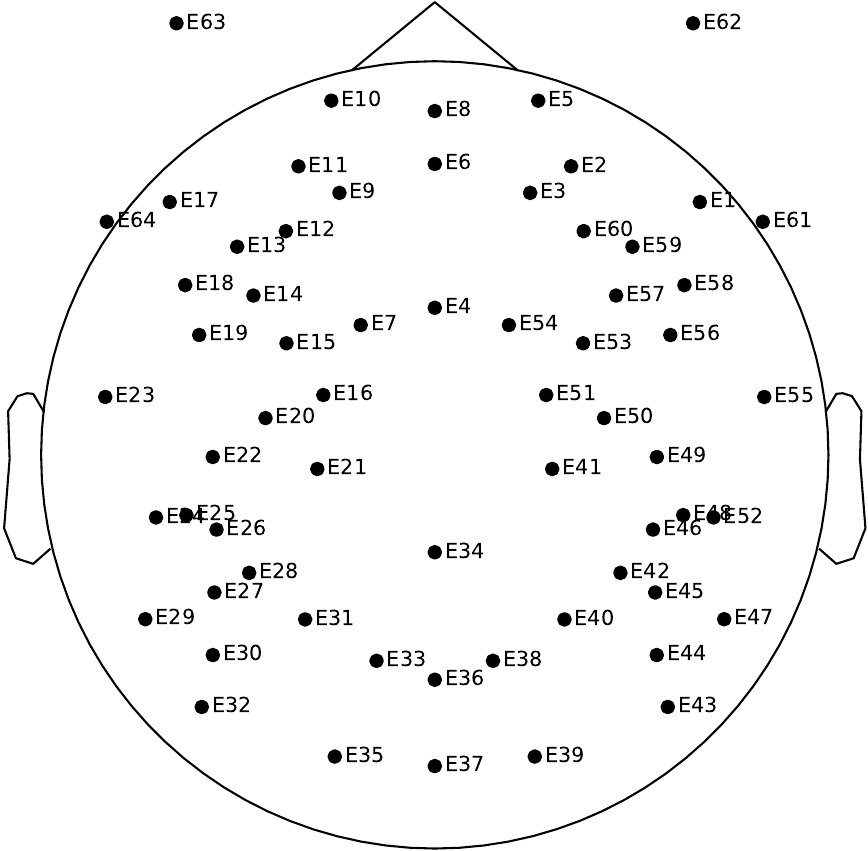}
\vskip -0.1in
\caption{
Electrode montage of a 64-channel EEG cap using the GSN-HydroCel-64\_1.0 layout. Sensor positions are annotated with their corresponding channel labels.
\label{fig:eeg_device}
}
\end{figure}

We illustrate the electrode montage of a 64-channel EEG cap configured according to the GSN-HydroCel-64\_1.0 layout in Fig.~\ref{fig:eeg_device}.

\section{Additional Experimental Results}
\label{sec:exps}

\subsection{Implementation Details}
Due to GPU memory constraints, Our proposed CineSync is trained using a two-stage approach. In the first stage, only the multimodal fusion encoder is trained using contrastive loss with a batch size of 16 on a single H100 GPU. To enhance the effectiveness of contrastive learning, we augment the training dataset with diverse textual captions generated by Qwen2.5-VL-7B~\cite{Qwen2.5-VL}. This pretraining phase lasts approximately 50 epochs. In the second stage, the pretrained encoder is integrated into the full model, which is finetuned for 5000 steps on 4 H100 GPUs using a batch size of 2 per GPU. The fMRI and EEG transformers each consist of 12 layers with a transformer dimension of 2048 and a token length of 227 (226 spatial tokens plus one class token). The LoRA configuration uses a rank of 64 and a scaling factor $\alpha$=64. Input data consist of 4-second multimodal brain signals, yielding standard inputs of 5$\times$8405 fMRI voxels and 64$\times$4000 EEG data points.

\begin{table*}
\small
\centering{
\caption{
\label{tab:align_ablation}
\textbf{Ablation study on multimodal alignment in CineSync.} 
We report the average performance across all subjects. 
CineSync$^{\star}$ indicates the experiment that includes audio-related ROIs in fMRI. 
}
\vskip -0.1in
    \setlength{\tabcolsep}{3.4mm}{
    \begin{tabular}{l|ccccc|cccc}
    \toprule
    \multicolumn{1}{c|}{\multirow{3}{*}{\textsc{Methods}}} & \multicolumn{5}{c|}{Semantic-level} & \multicolumn{4}{c}{Perceptual-level} \\
    & \multicolumn{3}{c|}{Video-based} &  \multicolumn{2}{c|}{Frame-based} &  \multicolumn{2}{c|}{Video-based} &  \multicolumn{2}{c}{Frame-based} \\
     & 2-way$\uparrow$ & 50-way$\uparrow$ & \multicolumn{1}{c|}{FVD$\downarrow$} & 2-way$\uparrow$  & 50-way$\uparrow$ & DTC$\uparrow$ & \multicolumn{1}{c|}{CTC$\uparrow$}  & SSIM$\uparrow$ & PSNR$\uparrow$ \\
    \midrule
    \midrule

\textit{w/o Vision}  & 0.863  & 0.275  & \multicolumn{1}{c|}{52.06} & 0.858 & 0.378 & 0.895 & 0.738  & 0.272  & 11.69 \\
\textit{w/o Text}    & 0.891  & 0.294  & \multicolumn{1}{c|}{50.15} & 0.887 & 0.394 & 0.908 & 0.949  & 0.285  & 11.95 \\ 
\textit{w/o Across}  & 0.873  & 0.279  & \multicolumn{1}{c|}{51.29} & 0.864 & 0.382 & 0.902 & 0.943  & 0.274  & 11.74 \\
\midrule
\textbf{CineSync$^{\star}$} &   \textbf{0.926}  & \textbf{0.336}  & \multicolumn{1}{c|}{\textbf{44.77}}  &  \textbf{0.954} & \textbf{0.423} & \textbf{0.921} & \textbf{0.953} & \textbf{0.297} & \textbf{12.18}  \\ 
    \bottomrule
    \end{tabular}}
    }
\vskip -0.05in
\end{table*}

\begin{figure*}
\centering
\includegraphics[width=\linewidth]{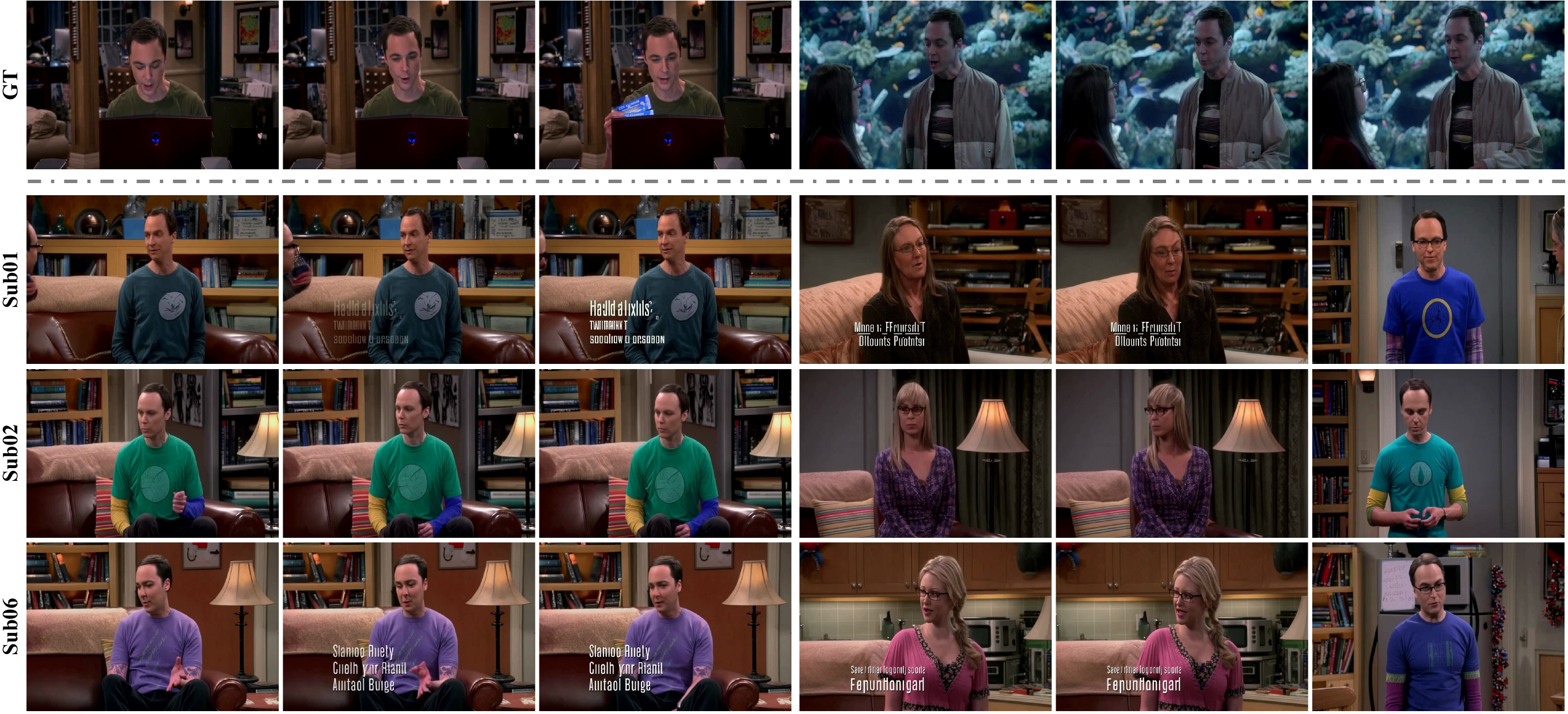}
\caption{ 
\textbf{Video Reconstruction Results for Subjects 1, 2, and 6.} 
We compare the reconstructed frames from Subjects 1, 2, and 6 with the corresponding ground-truth (GT) frames. The consistent semantic alignment and visual fidelity across subjects demonstrate the robustness and strong cross-subject generalization ability of \textbf{CineSync}.
\label{fig:video_across_sub_126}
}
\end{figure*}

\begin{figure*}
\centering
\includegraphics[width=\linewidth]{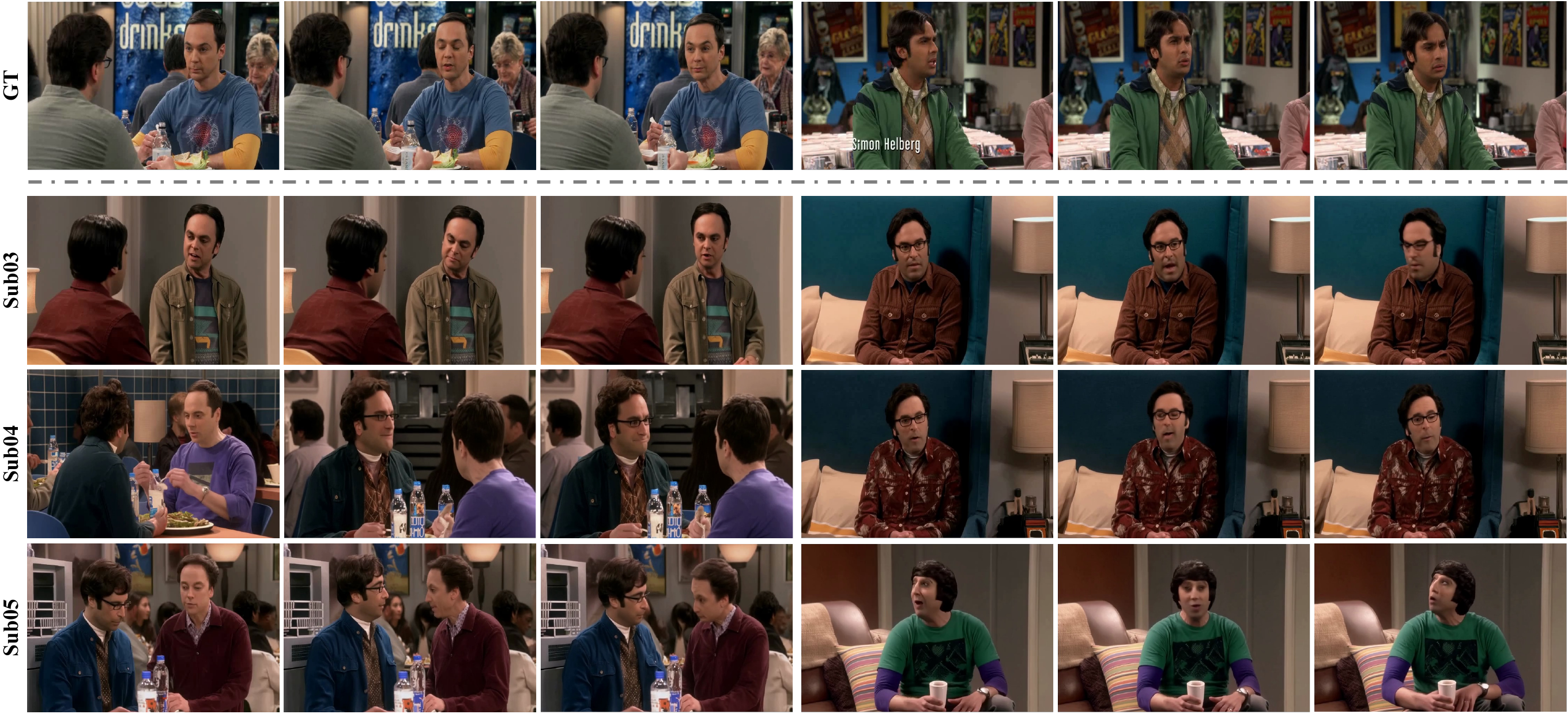}
\vskip -0.1in
\caption{
\textbf{Video Reconstruction Results for Subjects 3, 4, and 5.}
Reconstructed frames from Subjects 3, 4, and 5 are shown alongside the GT frames at matched timestamps. The results further verify that \textbf{CineSync} maintains stable reconstruction quality across different individuals.
\label{fig:video_across_sub_345}
}
\vskip -0.1in
\end{figure*}

\begin{table*}
\small
\centering{
\caption{
\label{tab:all_metric}
\textbf{Performance of each subject in CineBrain.} CineSync$^{\star}$ indicates the experiment that includes audio-related ROIs in fMRI.
}
\vskip -0.1in
    \setlength{\tabcolsep}{3.4mm}{
    \begin{tabular}{l|ccccc|cccc}
    \toprule
    \multicolumn{1}{c|}{\multirow{3}{*}{\textsc{Methods}}} & \multicolumn{5}{c|}{Semantic-level} & \multicolumn{4}{c}{Perceptual-level} \\
    & \multicolumn{3}{c|}{Video-based} &  \multicolumn{2}{c|}{Frame-based} &  \multicolumn{2}{c|}{Video-based} &  \multicolumn{2}{c}{Frame-based} \\
     & 2-way$\uparrow$ & 50-way$\uparrow$ & \multicolumn{1}{c|}{FVD$\downarrow$} & 2-way$\uparrow$  & 50-way$\uparrow$ & DTC$\uparrow$ & \multicolumn{1}{c|}{CTC$\uparrow$}  & SSIM$\uparrow$ & PSNR$\uparrow$ \\
    \midrule
    \midrule
Subject 1  & 0.921  & 0.332  & \multicolumn{1}{c|}{46.12} & 0.951 & 0.419 & 0.918 & 0.949 & 0.294 & 12.11 \\
Subject 2  & 0.928  & 0.337  & \multicolumn{1}{c|}{44.03} & 0.956 & 0.426 & 0.922 & 0.954 & 0.298 & 12.20 \\
Subject 3  & 0.923  & 0.334  & \multicolumn{1}{c|}{45.87} & 0.952 & 0.421 & 0.917 & 0.950 & 0.295 & 12.14 \\
Subject 4  & 0.927  & 0.338  & \multicolumn{1}{c|}{43.95} & 0.957 & 0.427 & 0.923 & 0.956 & 0.299 & 12.23 \\
Subject 5  & 0.929  & 0.339  & \multicolumn{1}{c|}{44.62} & 0.955 & 0.425 & 0.920 & 0.952 & 0.296 & 12.17 \\
Subject 6  & 0.924  & 0.335  & \multicolumn{1}{c|}{44.97} & 0.953 & 0.420 & 0.919 & 0.951 & 0.297 & 12.22 \\
\midrule
\textbf{CineSync$^{\star}$} &   \textbf{0.926}  & \textbf{0.336}  & \multicolumn{1}{c|}{\textbf{44.77}}  &  \textbf{0.954} & \textbf{0.423} & \textbf{0.921} & \textbf{0.953} & \textbf{0.297} & \textbf{12.18}  \\ 
    \bottomrule
    \end{tabular}}
    }
\end{table*}

\begin{figure*}
\centering
\includegraphics[width=\linewidth]{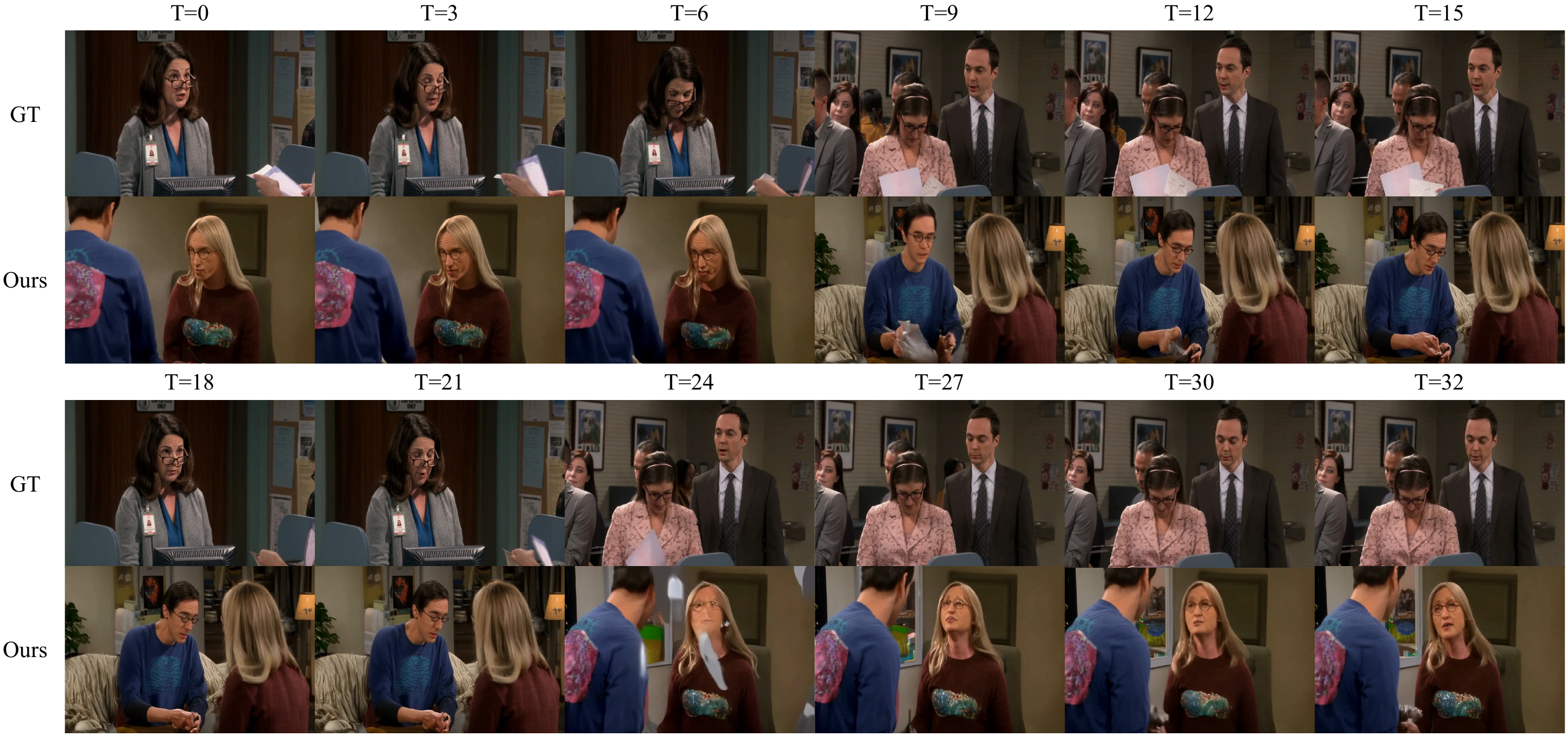}
\vskip -0.1in
\caption{ 
\textbf{More Results of CineSync:} We present 12 frames with timestamps compared with the ground truth (GT).
\label{fig:morevideo1}
}
\vskip -0.15in
\end{figure*}

\begin{figure*}
\centering
\includegraphics[width=\linewidth]{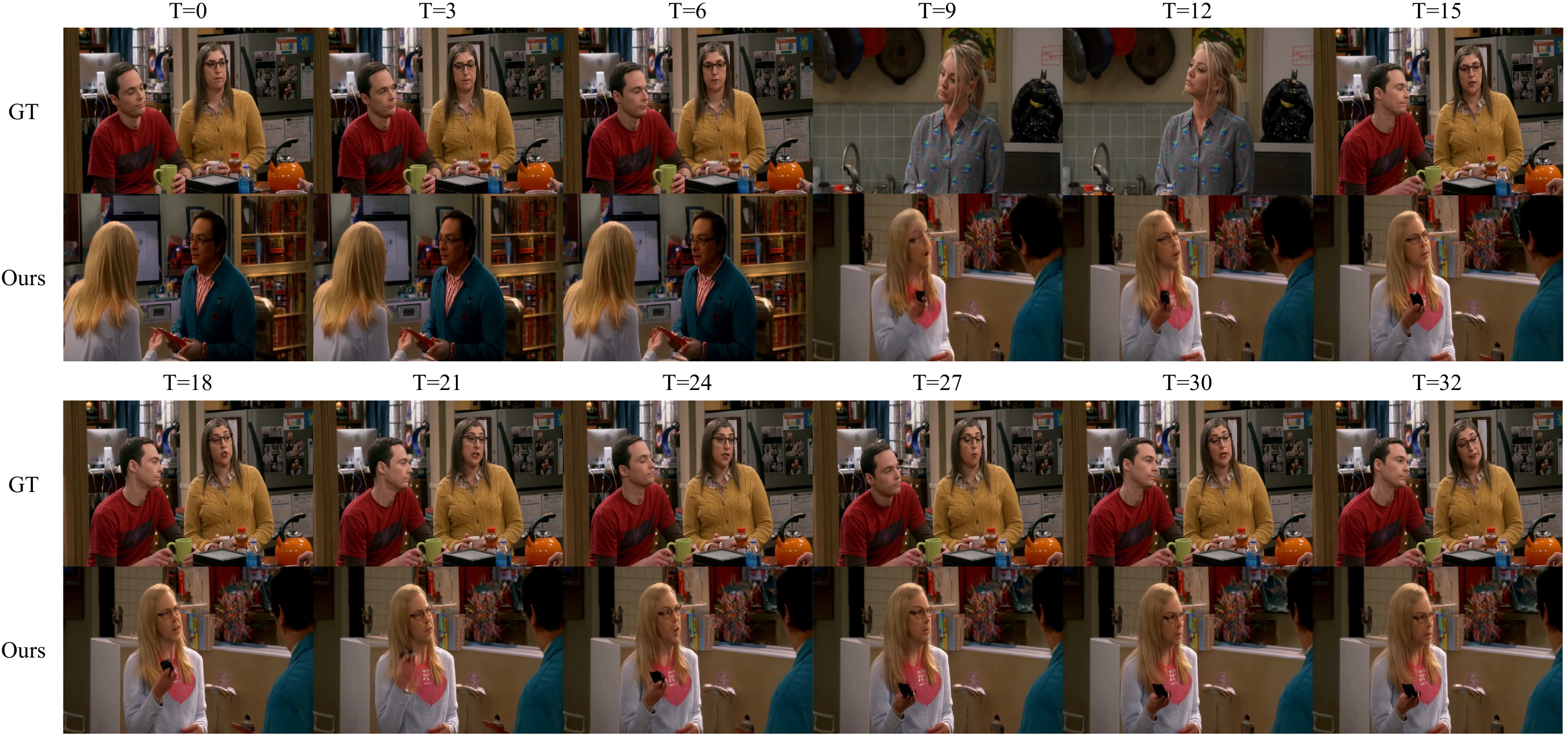}
\vskip -0.1in
\caption{ 
\textbf{More Results of CineSync:} We present 12 frames with timestamps compared with the ground truth (GT).
\label{fig:morevideo2}
}
\vskip -0.15in
\end{figure*}

\begin{figure*}
\centering
\includegraphics[width=\linewidth]{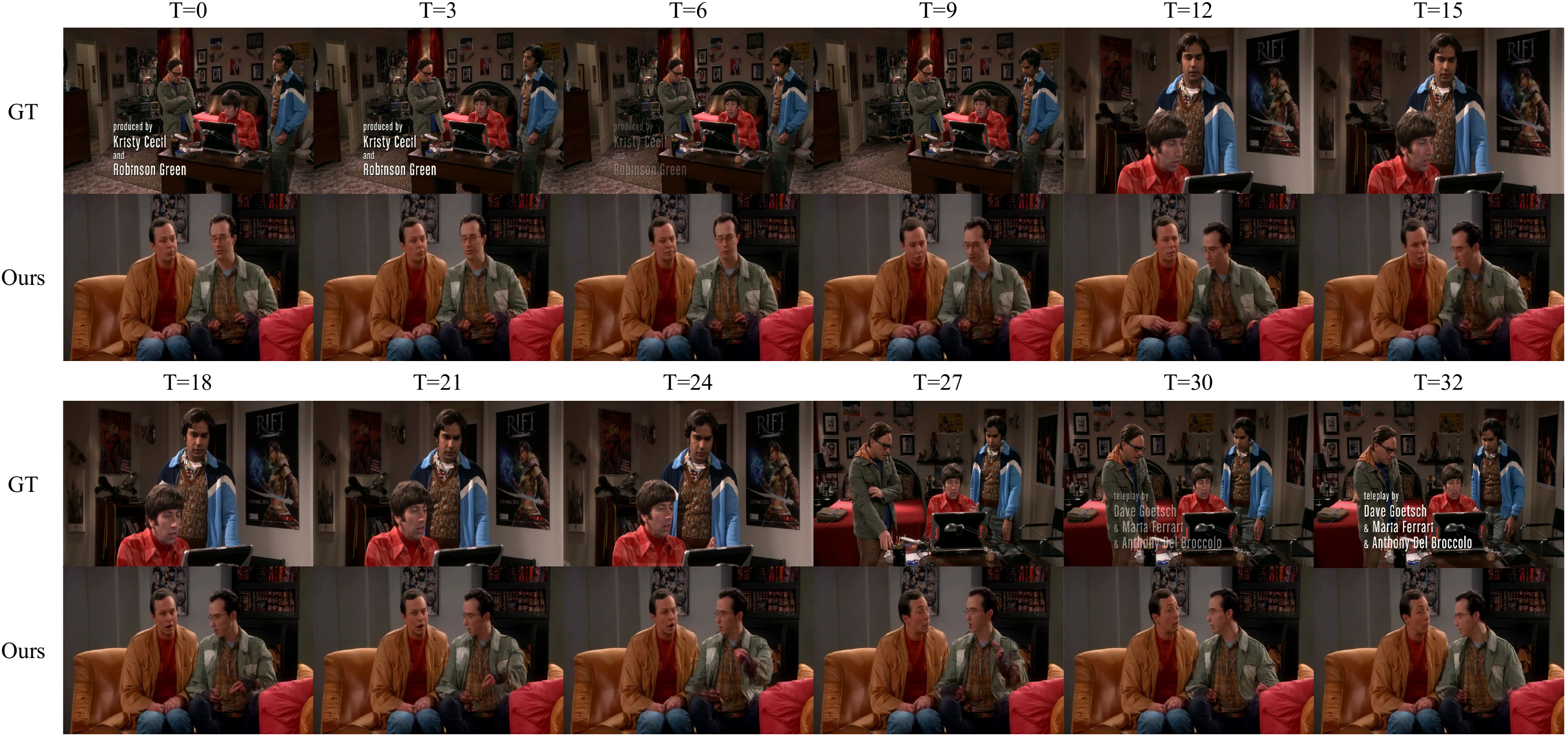}
\vskip -0.1in
\caption{ 
\textbf{More Results of CineSync:} We present 12 frames with timestamps compared with the ground truth (GT).
\label{fig:morevideo3}
}
\vskip -0.15in
\end{figure*}

\subsection{Ablation Study on Multimodal Alignment}
To quantify the contribution of each alignment component in \textbf{CineSync}, we conduct ablations on the full model trained with both audio- and vision-related ROIs in fMRI. Specifically, \textit{w/o Vision}, \textit{w/o Text}, and \textit{w/o Across} correspond to removing (i) the vision–fMRI alignment branch, (ii) the text–fMRI alignment branch, and (iii) the cross-modal alignment between EEG and fMRI, respectively. 
The averaged results across all subjects are reported in Tab.~\ref{tab:align_ablation}. Removing any of these alignment pathways leads to clear and consistent performance drops across both semantic- and perceptual-level metrics, demonstrating that rich cross-modal alignment is critical for achieving high-quality brain-to-video reconstruction.

\subsection{Detailed Results for Each Subject}
To further verify that our model remains robust across individuals, we report per-subject quantitative performance in Tab.~\ref{tab:all_metric}. As shown, the results exhibit only minor variations across the six subjects, indicating that \textbf{CineSync} generalizes well to different brain patterns. Moreover, the averaged scores closely match the overall CineSync$^{\star}$ results, further confirming subject-invariant performance.

\subsection{Video Reconstruction Across Subjects}
To better illustrate the robustness of our model, we visualize the reconstructed videos from different subjects under the same visual stimuli. Representative results are shown in Fig.~\ref{fig:video_across_sub_126} and Fig.~\ref{fig:video_across_sub_345}. Since Subjects 1, 2, and 6 share the same train–test split, and Subjects 3, 4, and 5 share another split, we compare the reconstruction quality within each group accordingly. Specifically, Fig.~\ref{fig:video_across_sub_126} presents the reconstructions of Subjects 1, 2, and 6, while Fig.~\ref{fig:video_across_sub_345} shows those from Subjects 3, 4, and 5. These results indicate that our model achieves stable performance across different individuals.

\subsection{More Results of Video Reconstruction}
To further showcase the quality, semantic fidelity, and temporal coherence of the reconstructed videos, we present additional 12-frame examples in Fig.~\ref{fig:morevideo1}, Fig.~\ref{fig:morevideo2}, and Fig.~\ref{fig:morevideo3}. Across diverse scenes, our method consistently produces reconstructions that are semantically aligned with the stimuli and temporally smooth.

\section{Limitations and Future Work}
Although our CineSync successfully leverages the spatial resolution of EEG to compensate for the temporal resolution limitations of fMRI, substantially improving video and audio reconstruction performance, our dataset currently does not explicitly provide additional fMRI data. This limitation restricts broader applications of our dataset and represents an area for future exploration.

Furthermore, our CineBrain dataset supports synchronized audiovisual stimuli. However, we have independently evaluated our primary contributions solely through separate video and audio reconstruction tasks. If we aim to expand into more complex applications such as embodied intelligence, it will be necessary to reconstruct multiple modalities simultaneously. Therefore, joint audiovisual reconstruction represents another significant direction for our future research.


\end{document}